\documentclass{article}

\usepackage[utf8]{inputenc} % allow utf-8 input
\usepackage[T1]{fontenc}    % use 8-bit T1 fonts
\usepackage{hyperref}       % hyperlinks
\usepackage{url}            % simple URL typesetting
\usepackage{booktabs}       % professional-quality tables
\usepackage{amsfonts}       % blackboard math symbols
\usepackage{nicefrac}       % compact symbols for 1/2, etc.
\usepackage{microtype}      % microtypography
\usepackage{xcolor}         % colors

\usepackage[final]{neurips_2021}
\usepackage{graphicx}
\usepackage{subfigure}
\usepackage{amsmath}
\usepackage{algorithm}
\usepackage{adjustbox}
\usepackage[toc,page]{appendix}
\usepackage{bbm}
\usepackage[noend]{algpseudocode}
\usepackage{lipsum} % for Lorem ipsum
\makeatletter
\def\algbackskip{\hskip-\ALG@thistlm}
\makeatother
\algrenewcommand\algorithmicrequire{\textbf{Input:}}
\algrenewcommand\algorithmicensure{\textbf{Output:}}
\usepackage{amssymb}
\usepackage{newunicodechar}
\usepackage[utf8]{inputenc}
\usepackage{booktabs} % for professional tables
\usepackage{hyperref}
\usepackage[skip=0pt]{caption}
\title{Adaptive Aggregation for Safety-Critical Control\vspace{-2em}}

\vspace{-5mm}
\author{%
  Huiliang Zhang, Di Wu and Benoit Boulet \\
  Department of Electrical \& Computer Engineering\\
  McGill University\\
  Montreal, QC H3A 0E9 \\
  \texttt{huiliang.zhang2@mail.mcgill.ca} \\
  \vspace{-6mm}
}
\vspace{-5mm}
\begin{document}
\maketitle
\begin{abstract}

% \textcolor{blue}{safty is the ...in rl, some research..but no one focus on } The unconstrained and suboptimal behaviours of agents may lead to costly negative consequences.
%Safety is the central obstacle preventing the use of reinforcement learning (RL) in real-world applications. The learning efficiency and performance in safe RL are hundred by the limited exploration within safety constraints. Current research focuses rarely on safe learning by exploiting knowledge from the experienced tasks, leading to the low learning efficiency when applying the safe RL in a new environment.In this work, we propose an adaptive aggregation framework for safety-critical RL settings, which aims to analogize the behaviour of humans solving a real-world problem. Our method comprises two key techniques to improve learning efficiency and safety performance. 1) We learn to transfer the safety knowledge by aggregating the multiple source tasks and a target task through the attention network, which can learn an optimal policy with no assumptions about the existence of optimal source tasks.2) We separate the goal of improving task performance and constraint satisfaction by utilizing a safeguard, which ensures the non-safety violations while maintaining high accumulated rewards in the learning process.  Experiments demonstrate that our algorithm can achieve competitive rewards with higher learning efficiency and fewer safety violations in several control tasks.
Safety has been recognized as the central obstacle to preventing the use of reinforcement learning (RL) for real-world applications. Different methods have been developed to deal with safety concerns in RL. However, learning reliable RL-based solutions usually require a large number of interactions with the environment. Likewise, how to improve the learning efficiency, specifically, how to utilize transfer learning for safe reinforcement learning, has not been well studied. In this work, we propose an adaptive aggregation framework for safety-critical control. Our method comprises two key techniques: 1) we learn to transfer the safety knowledge by aggregating the multiple source tasks and a target task through the attention network; 2) we separate the goal of improving task performance and reducing constraint violations by utilizing a safeguard. Experiment results demonstrate that our algorithm can achieve fewer safety violations while showing better data efficiency compared with several baselines.

% \textcolor{blue}{
% 1:add 2022 3 papers
% 2:call back introduction, modify the experiments according to contribution
% 3:inspired by multipolar A2T  advrl 
% 4:add baseline multipolar as baseline 
% 5:more experiments correlated with contribution, from safe RL.}

%Transferring knowledge from prior source tasks in solving a new target task can be useful in several learning applications.
% aggregate the safe actions provided by the source policies adaptively to maximize the target task performance. 2) Meanwhile, we learn an auxiliary network that predicts residuals around the aggregated safe actions, which ensures the target policy’s expressiveness even when some of the source policies perform poorly. 3)We separate the constraints and explorations and use an advantaged safeguard policy to ensure safety during the learning process.
\end{abstract}

\vspace{-5mm}
\section{Introduction}
\vspace{-1mm}
\label{sec_intro}
Reinforcement learning (RL) is a key technique to build autonomous agents which can learn and adapt to the changes of environments. Recent advances in RL have led to rapid progress in domains such as Atari \cite{mnih2015human}, Go \cite{silver2017mastering}, manipulation  \cite{nagabandi2020deep,sun2022fully} , locomotion tasks \cite{haarnoja2018soft,li2021reinforcement}, and business \cite{zhang2021bcorle,ma2021hierarchical}.  However, deploying RL algorithms to real-world applications faces a hurdle with safety concerns. When venturing into new regions of the state space during the unconstrained exploration, the agent may cause unaccepted failures, such as unfavourable impacts to people, property and the agent itself \cite{garcia2015comprehensive, chen2021context, SafeNearFuture2021,saboo2021reinforcement}. Moreover, the safety constraints may also limit the agents' ability to explore the entire state and action space to maximize the expected total reward \cite{thananjeyan2021recovery}. Thus, achieving adaptability and maintaining good performance with constraints satisfaction is of importance for the widespread use of RL in the real world.

% \textcolor{blue}{exaple on inefficiency of safe RL}
Most RL research endows agents with the ability to satisfy safety constraints from the line of control theory-based method or the constrained policy optimization formulation \cite{chow2018lyapunov,thananjeyan2021recovery}. 
Remarkably, those safe RL algorithms succeeded with surprisingly little access to prior knowledge about the experienced tasks. Though the ability to learn with minimal prior knowledge is desirable, it can lead to computationally intensive learning and limited exploration. 
% This learning inefficiency should be \textcolor{red}{contrasted with human behaviour.} When trying to master a new skill with safety constraints, our learning progress relies heavily on prior safety knowledge, which we have collected in solving previous instances of similar problems. 
Moreover, the control theory-based method learns a conservative safe region with the accurate dynamic model, which can remain no safety violations \cite{chow2018lyapunov,lutjens2019safe,cheng2019end,brown2020safe, SafeNearFuture2021,paternain2022safe}. The constrained policy optimization uses an intervention mechanism to evaluate the safety or by adding the penalty in reward functions to suppress the unsafe policy \cite{alshiekh2018safe,thananjeyan2021recovery,cowen2022samba}. Those two methods maintain good safety performance after the policy converged but may fail to work well in a new safety-critical settings \cite{zhang2020cautious,laroche2019safe,chen2021context,satija2021multiobjective}. 
%example
% When trying to master a new skill with safety constraints, our learning progress relies heavily on prior safety knowledge, which we have collected in solving previous instances of similar problems. 
% For example, when we learn how to drive a truck safely, we will recall our experience to drive a car to avoid pedestrians if it's possible, or the experience to drive in extreme weather like snowy and foggy. 

% \textcolor{blue}{intro transfer learningL}
% In a safe RL, the agents can develop the ability to fast adapt and quickly learn in a new environment setting if they leverage the previously trained policy. 
Transferring safety knowledge gained from tasks solved earlier to solve a new target safety-critical task can help, either in terms of speeding up the learning process or in terms of achieving a better performance \cite{laroche2019safe,zhang2020cautious,turchetta2020safe,chen2021context}.  The existing transfer RL approaches such as  \cite{rajendran2015attend} (A2T) and  \cite{barekatain2019multipolar} (MULTIPOLAR) omit the safety requirements which could lead to costly failure. A2T also fails to deal with partially useful policies and MULTIPOLAR can't attend to the changes of input states directly. Plus, safe policy reuse methods assume an optimal safe policy and focus on selecting a suitable source policy for explorations. They are
also unable to handle cases when the source policy is only partially useful for learning the target task \cite{zhang2020cautious,turchetta2020safe} (CARL and CISR). Although some transfer approaches have utilized multiple source policies during the target task learning, they have strong assumptions on the guaranteed relatedness
%severely limit the applicability of many transfer techniques tasks can be guaranteed beforehand
between source and target tasks \cite{turchetta2020safe,chen2021context}. Moreover, we cannot rely on a history of their individual experiences, as they may be unavailable due to a lack of communication between factories or prohibitively large dataset sizes \cite{chen2021context,laroche2019safe, turchetta2020safe,thananjeyan2021recovery}, and cannot assure the non-safety violations during the training and deployment. 
% For safety concerns, the current method could also not ensure the none safety violations because exploring the environment to learn about constraints requires a significant amount of constraints violations during learning \cite{zhang2020cautious, chen2021context}. 

To tackle the aforementioned challenges, we propose to use transfer learning to improve the learning efficiency in safe RL. Specifically, inspired by \cite{rajendran2015attend, barekatain2019multipolar}, we propose an adaptive aggregation architecture in safety-critical (AASC) control which reuses knowledge from multiple sources solutions.
% To solve the above problems, we develop an adaptive aggregation architecture in safety-critical (AASC) control. Our framework is a real-world analogy to improve learning efficiency and safety performance, considering multiple solutions that have experience in the correlated safe RL tasks. 
Our key idea is twofold; 1) The safety knowledge transfer by aggregating multiple source tasks and a target task through attention and auxiliary network.
By learning aggregation parameters to maximize the expected return at a target environment instance, we can adapt quickly thus improving the learning efficiency in unseen target tasks without knowing source environmental dynamics or source policy performances. Plus, the agent can decide which source solution to attend or suppress to, or to choose the solution from the auxiliary when the source tasks are irrelevant to solving the target task.
%, for a given input state. 
%pick and use the solutions from the auxiliary network when the source task solutions are irrelevant for solving the target task. 
2) We separate the goals of improving task performance and constraint satisfaction by utilizing a safeguard to improve safety performance. This separation allows the learned task policy to purely focus on collecting the most informative experiences and can maintain safety during training and deployment.
%
% Under very mild assumptions on the source policies and the safety of the backup policy used during the safeguard evaluation, we prove that running AASC with any RL algorithm can safely learn a policy that has good performance in the safety-critical settings. 
We also empirically validate AASC by comparing its performance with several standard transfer RL and safe RL algorithms in simulated control tasks. Our experimental results demonstrate the significant improvement in learning efficiency and safety performance with the proposed approach. We also conducted a detailed analysis of factors that affect the performance of AASC, and demonstrate that the AASC is an effective and generic framework for safe RL.
%our approach and found it Moreover, we investigate the factors that 
%worked well  even when  some of  the  source  policies  performed poorly in their original environment instance.

\vspace{-2mm}
\section{Preliminaries}
\vspace{-2mm}
\subsection{Safe reinforcement learning}
\vspace{-1mm}
We consider safe RL in a $\gamma$-discounted infinite horizon MDP $M$. An MDP can be expressed as a tuple $M = \left<S,A,P,R,\gamma\right>$, where $S$, $A$, $P$ and $R$ are the sets of states $s_t$, actions $a_t$, state transition probabilities $p$ and rewards $r$; $\gamma\in[0,1]$ is a discount factor accounting for future rewards. 
% A policy $\pi$ on $M$ is a mapping $\pi:S\rightarrow\triangle(A)$, where $\triangle(A)$ denotes probability distributions on $A$. 
A policy $\pi$ induces a trajectory distribution. 
% $\rho^{\pi}(\xi)$, where $\xi=(s_0,a_0,s_1,a_1,\dots)$ denotes a random trajectory. 
For a state distribution $d\in\triangle(S)$ and function $f:s\rightarrow\mathbb{R}$, we define $f(d)=\mathbb{E}_{s\sim d}[f(s)]$.  The initial state distribution of $\pi$ is $d_0^{\pi}$ and the average state distribution induced by $\pi$ is $d^{\pi}=(1-\gamma)\sum_{t=0}^{\infty}\gamma^{t}d_{t}^{\pi}(s)$. The state-action value function of $\pi$ is defined as 
$Q^{\pi}(s, a) = \mathbb{E}_{\xi\sim\rho^{\pi}|s_0=s, a_0=a}[\sum_{t=0}^{\infty}\gamma^{t}r(s_t, a_t)]$ and its state value function as $V^{\pi}(s)=Q^{\pi}(s,\pi)$. The optimal stationary policy of $M$ is $\pi^{*}$ and its respective value function are $Q^{*}$ and $V^{*}$. 
% The random variable in the subscript of expectations is omitted if it is clear. 

The definition of safety is that the probability of the agent entering an unsafe subset of state $S_{unsafe} \subset\ S$ is low, which is consistent with \cite{SafeNearFuture2021,thananjeyan2021recovery,wagener2021safe}. We assume that we know the unsafe subset $S_{unsafe}$ and the safe subset $S_{safe} = S \setminus S_{unsafe}$. However, we make no assumption on the knowledge of reward $r$ and dynamics $P$, except that the reward $r$ is zero on $S_{unsafe}$ and that $S_{unsafe}$ is absorbing: once the agent enters $S_{unsafe}$ in a rollout, it cannot travel back to $S_{safe}$ and stays in $S_{unsafe}$ for the rest of the rollout. 
%
% \textcolor{blue}{A major concern toward safe RL is the existence of meta states $S_{\triangleright}$ which are currently safe but will lead to unsafe states with unsafe future actions. We denote that a state $s$ is viable if there exists a policy $\pi$ such that $\pi$ is safe starting from $s$, that is, executing $\pi$ starting from $s$ for infinite steps never leads to an unsafe state. And the safe state that is not viable is called a meta state and will have a chance to lead to an unsafe state. When an agent is going to violate constraints, 
% it could directly go to $S_{unsafe}$ or 
% it first enters $S_{\triangleright}$, and then it goes to the absorbing state $S_{unsafe}$ and stays there forever. }
%
Our goal is to find a policy $\pi$ that is safe and has a high return in $M$, and to do so via a safe data collection process, which is shown as follows: 
% Specifically, while keeping the agent safe during exploration, we want to solve the following chance constrained policy optimization problem:
\begin{equation}
\label{eq:basic formulation}
    \pi^{*}=\operatorname*{argmax}_{\pi}\{V^{\pi}(d_0):(1-\gamma)\sum_{h=0}^{\infty}\gamma^{h}\operatorname{Prob}(\xi_{h}\subset S_{safe}|\pi)\geq 1-\delta\}
\end{equation}
where  $\xi_h=(s_0,a_0, s_1,a_1,\dots,s_{h-1},a_{h-1})$ denotes an $h$-step trajectory segment and $\delta\in[0,1]$ is the tolerated failure probability.  $\operatorname{Prob}(\xi_h\subset S_{safe}|\pi)$ denotes the probability of $\xi_h$ being safe (i.e., not entering absorbing state from time step $0$ to $h-1$) under the trajectory distribution $\rho^\pi$ of $\pi$ on $M$. 
% A trajectory is safe if and only if all the states in the trajectory are safe. 
An initial state drawn from $d_0$ is assumed to safe with probability 1. %We say a deterministic policy $\pi$ is safe starting from state $d_0$, if the infinite-horizon trajectory obtained by executing $\pi$ starting from $d_0$ is safe. We also say a policy $\pi$ is safe if it is safe starting from an initial state drawn from $d_0$ with probability 1. 
The constraint shown in equation (1) is known as a chance constraint. The definition here accords to an exponentially weighted average (based on the discount factor $\gamma$) of trajectory safety probabilities of different horizons. Then the problem in equation (1) can be formulated as a constrained MDP (CMDP) problem with extra constraint cost function $C: S \times A \rightarrow \{0,1\}$ with associated discount factor $\gamma_{risk}$  which indicates whether state action is constraint violating. This yields the following new CMDP: $\widetilde{M}=<S, A, P, R, \gamma, C, \gamma_{risk}>$. And $\widetilde{M}$ consists with $M$ and a cost-based MDP  ($\overline{M}=<S, A, P, C, \gamma_{risk}>$). The chance-constrained policy optimization in (\ref{eq:basic formulation}) corresponds to the CMDP formulation from \cite{DBLP:journals/corr/abs-2003-02189} can be written as:
%where $c$ is the cost $c(s,a)=\mathbbm{1}\{{s \in (S_{\triangleright} \bigcup S_{unsafe})}\}$. 
\begin{equation}
\label{eq:cmdp formulation}
    \pi^{*}=\operatorname*{argmax}_{\pi}\{V^{\pi}(d_0):\overline{V}^{\pi}(d_0)\leq\delta\}
\end{equation}
where $\overline{V}^{\pi}(s)=\overline{Q}^{\pi}(s,\pi)$ and $\overline{Q}^{\pi}(s, a) = \mathbb{E}_{\xi\sim\rho^{\pi}|s_0=s, a_0=a}[\sum_{t=0}^{\infty}\gamma_{risk}^{t}c(s_t, a_t)]$. Equation (2) aims to find a policy that has a high cumulative reward $V^{\pi}(d_0)$ with cumulative cost $\overline{V}^{\pi}(d_0)$ below the allowed failure probability $\delta$. We assume that episodes terminate on violations, equivalent to transitioning to a constraint-satisfying absorbing state with zero reward. 
% The proof of this equivalence can be find in appendix \ref{sec:mdp proof }. 
% This setting exactly corresponds to the CMDP formulation from \cite{DBLP:journals/corr/abs-2003-02189}, but with constraint costs limited to binary indicator functions for constraint violating states. We limit the choice of the cost function to binary indicator functions, as they are easier to provide than shaped costs and use $\overline{Q}$ to convey information about delayed constraint costs. 
% Current CMDP formulation has been commonly studied to design safe RL algorithms to find good policies, such as the control theory-based method or the constrained policy optimization formulation. They learn a conservative safe region with the accurate dynamic model or add the penalty to suppress the unsafe policy \cite{laroche2019safe,zhang2020cautious,turchetta2020safe,chen2021context}.
% However, these algorithms rarely utilize prior knowledge about the experienced tasks, which can lead to computationally intensive learning, especially in safe RL settings where the exploration is limited. Moreover, they do not necessarily ensure safety during training and can be numerically unstable when we want to use them in a new safety-critical RL setting. 
% At a high level, this instability stems from the lack of off-the-shelf computationally reliable and efficient solvers for large-scale constrained stochastic optimization.
\subsection{Transfer reinforcement learning}
\vspace{-1mm}
Transfer RL aims at improving the  learning  efficiency  of  an  agent  by exploiting  knowledge  from  other  agents trained  on  source tasks \cite{barekatain2019multipolar}. 
Source tasks refer to tasks that we have already learnt to perform and target task refers to the task that we are interested in learning now .  
% \textcolor{red}{You should explain what are the source task and target task}
Here the source tasks should be in the same domain as the target task, having the same state and action spaces. Let there be $I$ source tasks which correspond to $I$ instances of the same environment which differ only in their state transition dynamics. Namely, we model each environment instance by an indexed MDP: $M_i= <S,A,P_i,R,\gamma>$ where no two-state transition distributions $P_i$, $P_j, i\neq j$ are identical. We assume that each $P_i$ is unknown when training a target policy, i.e., agents cannot access the exact form of $P_i$ nor a collection of states sampled from $P_i$. For each of the $I$ environment instances, we are given a source policy solution $\pi_{src}^{i}: S \rightarrow \triangle(A)$ that only maps states to actions. These solutions could be for example policies or state-action values.

\begin{figure}[t]
  \centering
  \includegraphics[width=12cm]{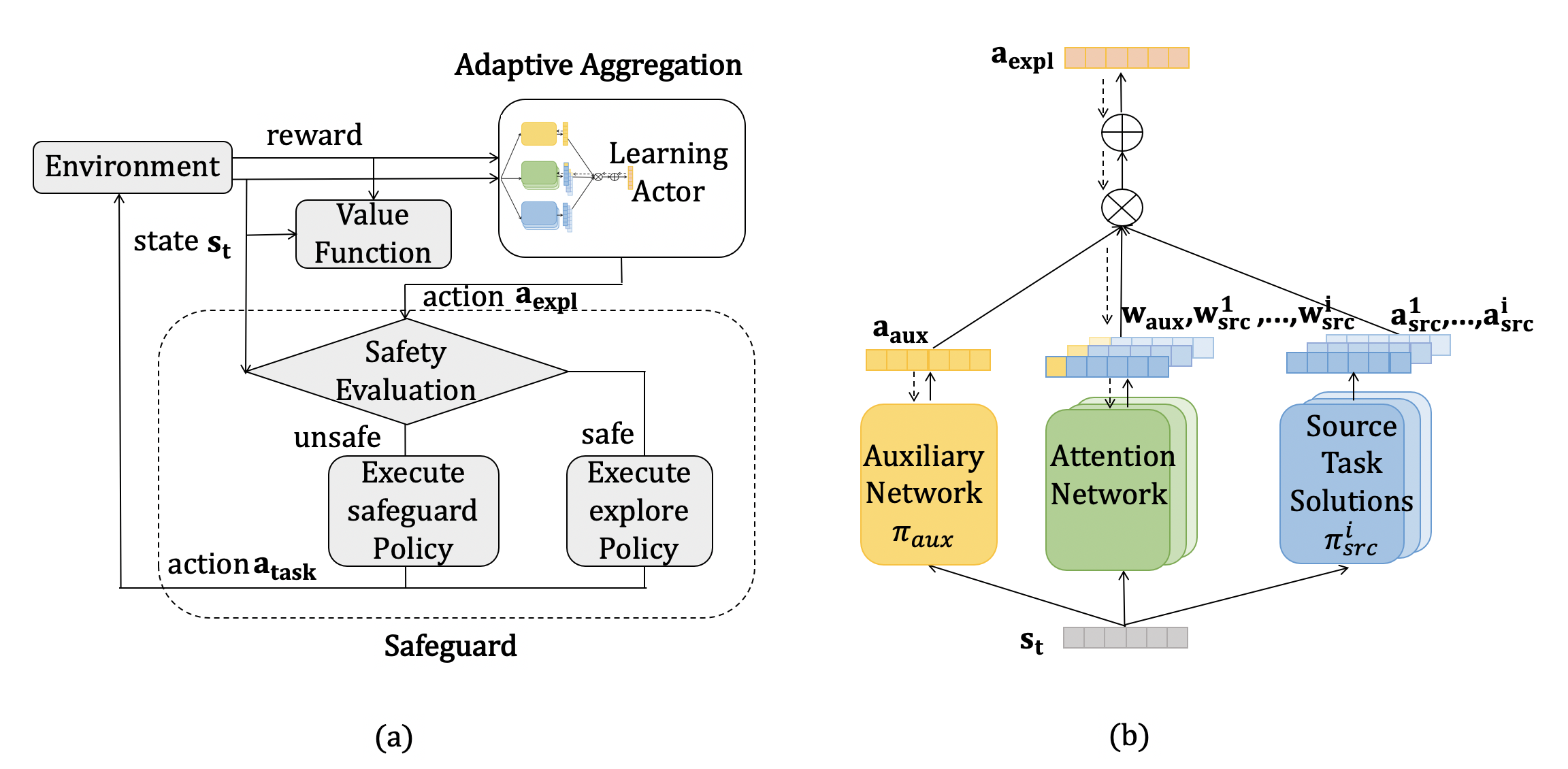}
  \label{fig_com}
  \caption{(a) Overview of AASC architecture. The adaptive aggregation module learns the exploratory action $a_{expl}$ and the safeguard module ensures the safety during the learning process and output task safe action $a_{task}$. (b) Adaptive aggregation of safe policies. We formulate the aggregated policy $\pi_{expl}$ with the sum of 1) the adaptive aggregation of  actions from source policies $\pi_{src}$ and 2) the auxiliary network $\pi_{aux}$ for predicting residuals $a_{aux}$. The dotted arrows represent the path of back propagation.}
\end{figure}
\vspace{-2mm}

\section{Adaptive aggregation in safety-critical (AASC) control}
In this section, we elaborate on the framework of AASC in safe RL settings, including the adaptive aggregation of safety policies and the safeguard evaluation parts.
%
% To analogize human behaviour to solve a new real-world problem with safety constraints, we propose an AASC framework for safe RL settings. 
In safety-critical settings, humans tend to query the knowledge learned before solving similar problems and get a potential solution. Then they will evaluate the safety of the potential solution and decided whether to execute it in real life.  The solution with safety constraints satisfaction could be found quickly in this manner.  Thus, the AASC consists of two parts: \textbf{the adaptive aggregation of safety policies} and \textbf{the safeguard evaluation}, as shown in figure 1 (a). The former helps us to efficiently learn the safe policy of a target agent given a collection of source policies, which inspired by \cite{rajendran2015attend, barekatain2019multipolar}, the latter ensures the safety constraints and maximizes the final performance in the safety-critical settings,  inspired by \cite{bharadhwaj2020conservative, wagener2021safe}. 
\begin{algorithm}[b]
\caption{Adaptive Aggregation in Safety-Critical (AASC) Control}
\label{algo_ac}
\begin{algorithmic}[1]
\Require AASC RL algorithm $\mathcal{F}$, $\mathcal{D}_{task}\leftarrow \emptyset$, $\mathcal{D}_{Safeguard} \leftarrow \emptyset$,  multi-source tasks solutions $\pi_{src}$.
\Ensure Optimized safe policy $\hat{\pi}^{*}$
\State $\mathcal{F}.\mathrm{Initialize()}$
\State $s \leftarrow$ env.reset()
\While {training budget available}
        \State $ a_{expl} \leftarrow \mathcal{F}.\mathrm{SafePolicyAggregation}$$(s,\pi_{src})$
        \State $ a_{task} \leftarrow \mathcal{F}.\mathrm{Safeguard}$($s, a_{expl}$)
        \State Execute $a_{task}$ and collect data to $\mathcal{D}_{task}$ and $\mathcal{D}_{Safeguard}$, $s = s^{'}$
        \State $\hat{\pi}\leftarrow$$\mathcal{F}.\mathrm{OptimizePolicy}(\mathcal{D}_{task})$
        \State $\mathcal{F}$.$\mathrm{UpdateSafeguardRule}(\mathcal{D}_{Safeguard})$
\EndWhile
\State \textbf{end while}
\State $\hat{\pi}^{*}\leftarrow \mathcal{F}.\mathrm{GetOptimizePolicy()}$
\end{algorithmic}
\end{algorithm}

The proposed method optimizes policies iteratively as outlined in Algorithm 1. 
As input, it takes an AASC algorithm $\mathcal{F}$ with a safe policy aggregation and safeguard module and multi-source tasks solutions $\pi_{src}$. The RL algorithm  $\mathcal{F}$ finds a nearly optimal policy for the MDP $\widetilde{M}$ constructed by the safeguard together with $M$, which is an approximate solution of equation \eqref{eq:cmdp formulation}. 
During training, the agent can interact with the unknown MDP $M$ to collect data under a training budget, such as the maximum number of environment interactions or allowed unsafe trajectories the agent can generate. 
In every iteration, the proposed method first queries the safe policy aggregation of $\mathcal{F}$ to have a potential exploratory action  $a_{expl}$ to execute in $\widetilde{M}$. Then it uses a safeguard policy to modify exploratory action into task safe action $a_{task}$.  The safeguard module is a shielded policy such that the agent runs backup policy $\mu:S\rightarrow\triangle(A)$ instead of $\pi_{expl}$ when $\pi_{expl}$ proposed unsafe actions. Then running  $a_{task}$ in the $M$ can be safe with high probability. Next, it collects data by running  $a_{task}$ in $M$ and collects data into $\mathcal{D}_{task}$ then transforms it into data $\mathcal{D}_{Safeguard}$. The transition stores in $\mathcal{D}_{task}$ is $<s_t, a_{task}, s_{t+1}, r_t>$ and $<s_t, a_{expl}, s_{t+1}, r_t, c_t>$ in $\mathcal{D}_{Safeguard}$. It then feeds $\mathcal{D}_{task}$ to the $\mathcal{F}$ for policy optimization and uses $\mathcal{D}_{Safeguard}$ to refine the shield policy. The process above is repeated until the training budget is used up. When this happens, it terminates and returns the best policy $\hat{\pi}^{*}$ from algorithm $\mathcal{F}$ can produce.
% \vspace{-1mm}
\subsection{Adaptive aggregation of safe policies}
The goal of this adaptive aggregation of safe policies is to train a new target agent’s policy $\pi_{task}$ in a sample efficient fashion. The target agent interacts with the target environment instance which is not identical to the source due to their distinct dynamics. For each of the $I$ source tasks, we are given the source policy $\pi_{src}^{i}:S \rightarrow A$ that only maps states to actions. 
% Source tasks refer to tasks that we have already learnt to perform and target task refers to the task that we are interested in learning now. 
% Here the source tasks should be in the same domain as the target task, having the same state and action spaces.  
Each source policy $\pi_{src}^{i}$ can be either parameterized (e.g., learned by interacting with its environment instance $M_{i}$) or non-parameterized (e.g., heuristically designed by humans). Either way, we assume that no prior knowledge about the source policies $\pi_{src}^{i}$ is available for a target agent, such as their representations of original performances, except that they were acquired from a source environment instance with an unknown dynamics. 
As shown in figure 1 (b), with the adaptive aggregation of safety policies, a target policy is formulated with the adaptive aggregation of actions from the set of source policies, and the auxiliary network mimicking the selected policies and predicting residuals around the aggregated actions. Let $a_{src}^{1}, a_{src}^{2}, \dots, a_{src}^{I}$ be the solutions of these source tasks $1, \dots, I$ respectively. $a_{aux}$ is the solution of an auxiliary network that starts learning from scratch while acting on the target task. Let $a_{expl}$ be the solution that we learn in the target task. The action space $a_{t}^{i}\in \mathbb{R}^{D}$ is a $D$-dimensional real-valued vector representing $D$ actions performed jointly in each timestep. For the collection of source policies, we derive the matrix of their actions:
\begin{equation}
    A_t = [(a_{src}^{1})^{\mathrm{T}}, \dots, (a_{src}^{I})^{\mathrm{T}}, (a_{aux})^{\mathrm{T}}] \in \mathbb{R}^{(I+1)\times D}
\end{equation}
The key idea of the aggregation module is to aggregate $A_t$ adaptively in an RL loop, i.e., to maximize the expected return $V^{\pi}(d_0)$. The adaptive aggregation only contains the source policies action that could introduce a strong inductive bias in the training of a target policy. So we learn an auxiliary policy network as $\pi_{aux}: S \rightarrow  \triangle(A)$ jointly with the source task policy, to predict residuals around the aggregated source task actions. 
$\pi_{aux}$ is used to improve the target policy training in two ways. 1) If the aggregated actions from $\pi_{src}$ are already useful in the target environment instance, $a_{aux}$ will correct them for a higher expected return. 2) Otherwise,  $\pi_{aux}$ learns the target task while leveraging  $\pi_{aux}$ as a prior to have a guided exploration process. Any network could be used for $a_{aux}$ as long as it is parameterized and fully differentiable. 

While the source task solutions $a_{src}^{1}, a_{src}^{2}, \dots, a_{src}^{I}$ remain fixed, the auxiliary network solutions are learnt and hence $a_{aux}$ can change over time. There is an attention network to learn the weights $w_{aux}, w_{src}^{1}, w_{src}^{2}, \dots, w_{src}^{I}$ given the input state $s_t$. The weights determine the attention each actions gets, allowing the agent to selectively accept or reject the different actions, depending on the input states. The aggregation policy is formulated as:
\begin{equation}
   a_{expl} =  W_t \odot A_t
\end{equation}
where $W_t = [ w_{src}^{1}, w_{src}^{2}, \dots, w_{src}^{I}, w_{aux}] \in \mathbb{R}^{(I+1)\times D}$ is the weight matrix. $\odot$ is the element-wise multiplication.
$W_t$ is neither normalized nor regularized and can scale each action of each policy independently. In this way, we can flexibly emphasize informative source actions while suppressing irrelevant ones. If the $i$ source task solution's action is useful at state $s$, then the corresponding element in $w_{src}^{i}$ is set to a high value by the attention network. Working at the granularity of states allows the attention network to attend to different source tasks, for different parts of the state space of the target task, thus giving it the ability to select informative actions. For parts of the state space in the target task, where the source task solutions are not relevant or even perform badly, the attention network learns to give high weight to the auxiliary network solution (which can be learnt and improved), thus avoiding bad source actions. The adaptive safe policy aggregation is shown in algorithm 3.

Depending on the feedback obtained from the environment upon following $a_{expl}$, the attention network’s parameters are updated to improve performance. As mentioned earlier, the source task solutions, $a_{src}^{1}, a_{src}^{2}, \dots, a_{src}^{I}$ remain fixed. Updating these source task’s parameters would cause a significant amount of unlearning in the source tasks solutions and result in a weaker transfer, which we observed empirically. 
%This also enables the use of source task solutions, as long as we have the outputs alone, irrespective of how and where they come from. 
Even though the agent follows $a_{expl}$, we update the parameters of the auxiliary network that produces $a_{aux}$, as if the action taken by the agent was based only on $a_{aux}$. Due to this special way of updating $a_{aux}$, %apart from the experience got through the unique and individual contribution of $a_{aux}$ to $a_{expl}$ in parts of the state space where the source task solutions are not relevant, 
$a_{aux}$ also uses the valuable experience got by using $a_{expl}$ which uses the solutions of the source tasks as well. 
This also means that, if there is a source task whose solution $a_{src}^{i}$ is useful for the target task in some parts of its state space, then $a_{aux}$ tries to replicate $a_{src}^{i}$ in those parts of the state space. In practice, the source task solutions though useful might need to be modified to suit perfectly the target task. The auxiliary network takes care of these modifications required to make the useful source task solutions perfect for the target task. The special way of training the auxiliary network assists the architecture in achieving this faster. 
%Note that the agent could follow $a_{src}^{i}$ through $a_{expl}$ even when $a_{aux}$ does not attain its replication in the corresponding parts of the state space. 
% This allows for a good performance of the agent in earlier stages of training itself when a useful source task is available and identified.
\begin{algorithm}[t]
\caption{Adaptive Aggregation of Safety Policies}
\label{algo_agg}
\begin{algorithmic}[1]
\Require State $s$, multi-source tasks solutions $\pi_{src}$ 
\Ensure Adaptive aggregation action $a_{expl}$
    \For{$i \in \{1,...,I\}$}
        \State Calculate $a_{src}^{i} \sim \pi_{src}^{i}(\cdot|s)$
    \EndFor
    \State Calculate $a_{aux} \sim \pi_{aux}(\cdot|s)$
    \State Calculate $w_{aux}, w_{src}^{1}, w_{src}^{2}, \dots, w_{src}^{I} $
    \State Calculate $a_{expl}$ according to equation (4).
\end{algorithmic}
\end{algorithm}
\subsection{Safeguard evaluation}
The goal of safeguard evaluation is to evaluate the safety given the potential actions $a_{expl}$ under the given states, and to guide the agent to safety when there are constraint violations likely.
% Safe exploration is challenging because it is difficult to detect meta states with actions that could lead to violations. 
Most prior work in safe RL integrates constraint satisfaction into the task objective to jointly optimize the two and detect those regions.
However, the inherent objective conflict exploration and constraints can lead to suboptimzlities in policy optimization. In the proposed safeguard, we consider an RL formulation subject to constraints on the probability of unsafe future behaviour and design the algorithm that can balance the often conflicting objectives of task-directed exploration and safety, which is inspired by \cite{thananjeyan2021recovery, wagener2021safe}. The agent evaluates the safety of $a_{expl}$ in the safeguard module, and instead executes approximate resets to nearby safe states when constraint violation is probable. 
% The meta states and actions could be evaluated directly by the cost function. 
\begin{figure}[b]
  \centering
  \includegraphics[width=8cm]{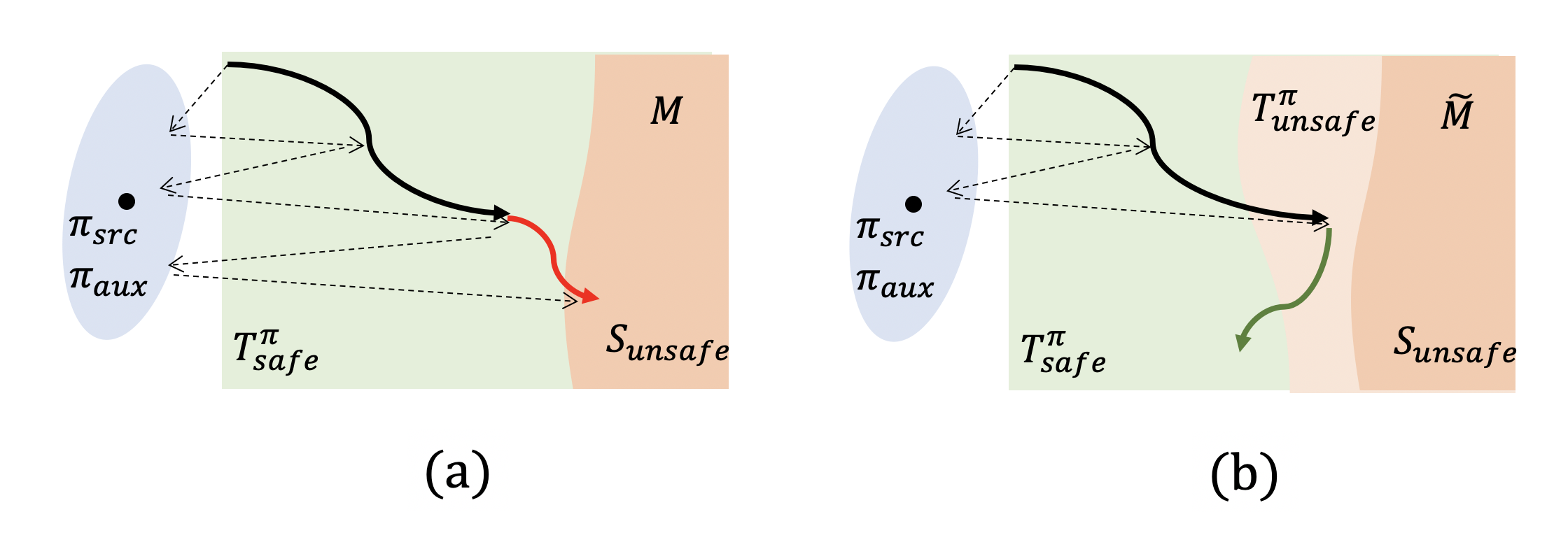}
  \label{fig_safe_ev}
  \caption{Safeguard Evaluation. (a) The agent in AASC starts in the safe states and follows the policy which is projected from the source and auxiliary policies. Without the safeguard evaluation, the agents may execute the disadvantaged action and have safety violations as red path. (b) Under the protection of safeguard, the backup policy will be activated and guide the agent to safety as green path.}
\end{figure}
 %The cost function is designed to address this policy $\pi$ certified by some $h$ guarantees to stay within $C_h$ and therefore can be used for collecting data. However, we may need more diversity in the collected data beyond what can be offered by the deterministic certified policy $\pi_{safeguard}$. Thanks to the contraction property R.3, we in fact know that any exploration policy $\pi_{expl}$ within $C_{h}$ can be made safe with $\pi_{safeguard}$ being a safeguarding policy, 
%we can first try actions \textbf{$\pi_{expl}$} and see if they stay within the viable subset $C_h$, and if none does, invoke the safeguarding policy $\pi_{safeguard}$. Algorithm 3 describes formally this simple procedure that makes any exploration actions $\a_{expl}$ safe. By a simple induction, one can see that the policy defined in Algorithm 3 maintains that all the visited states lie in $C_{h}$. 

To quantify the risk of entering an unsafe state, the safeguard rule is specified by a tuple $\mathcal{G}=<\overline{Q}, \mu, \eta>$, where $\overline{Q}: S \times A \rightarrow[0,1]$ is a state-action risk value estimator, $\eta$ is a threshold and $\mu$ is a backup safeguard action from $\pi_{backup}$ policy \cite{thananjeyan2021recovery}. $\pi_{backup}$ is supposed to safeguard the exploration. 
As shown in algorithm (\ref{algo:advshield}), when sampling $a_{task}$ from safeguard policy, it first queries if $(s, a_{expl})\in \mathcal{T}_{unsafe}^{\pi}$ then it samples  $a_{task}$ according to $\pi_{backup}$. Otherwise executes $a_{task} = a_{expl}$. However, activating the backup policy too often is undesirable, as it only collects data from $\pi_{backup}$ so there will be little exploration.
Hence we define the unsafe set $\mathcal{T_{\text{unsafe}}^{\pi}}$ in safeguard as: 
\begin{equation}
\label{t_unsafe}
\begin{split}
\mathcal{T}_{unsafe}^{\pi} &= \{(s,a)\in S_{safe}\times A:\overline{A}(s,a_{task})\geq\eta\} \\
\mathcal{T}_{safe}^{\pi}  &= S\times A  \setminus \mathcal{T}_{unsafe}^{\pi}   
\end{split}    
\end{equation}
and  advantage cost function is:
\begin{equation}
\label{advant}
\overline{A}(s,a)=\overline{Q}(s,a)-\overline{Q}(s,\mu)
\end{equation}
The equation (\ref{t_unsafe}) and (\ref{advant}) mean that we have assumption: for all $(s, a) \in \mathcal{T}_{unsafe}^{\pi} $ that can be reached from $d_0$ with some policy, there exist some $a \in A$ such that $\overline{A}(s, a) = \overline{Q}(s, a) -\overline{Q}(s, \mu) \geq \eta$. In other words, for every state action we can reach from $d_0$ that will be overridden, there is an alternative action in the agent’s action space $A$ that keeps the agent’s policy being safe.
By running the safeguard evaluation constructed by the advantage function $\overline{A}$, our method controls the safety relative to the backup policy $\mu$ concerning $d_0$. 
If the relative safety for each time step (i.e., advantage) is close to zero, then the relative safety overall is also close to zero (i.e. $\overline{V}^{\pi}(d_0)\leq \delta$).
Note that the safeguard while satisfying $\overline{V}^{\pi}(d_0)\leq \delta$, can generally visit (with low probability) the states where $\overline{V}^\mu(s)>0$ (e.g., $=1$). At these places where $\mu$ is useless for safety, the safeguard rule naturally deactivates and lets the learner explore, which avoids the overly conservative safe region in \cite{thananjeyan2021recovery,bharadhwaj2020conservative}. 
\begin{algorithm} [t]
\caption{Safeguard Evaluation}
\label{algo:advshield}
\begin{algorithmic}[1]
\Require State $s$, adaptive aggregation action $a_{expl}$
\Ensure  task safe action $a_{task}$
    \If{$s,a_{expl} \in \mathcal{T_{\text{unsafe}}^{\pi}}$}
    \State $a_{task} \sim {\pi_{backup}}(\cdot|s) $ 
    \Else
    \State $a_{task} = a_{expl}$
    \EndIf
\end{algorithmic}
\end{algorithm}
When the agent takes some actions $(s, a) \in \mathcal{T}_{unsafe}^{\pi}$ in $\widetilde{M}$, it goes to an  absorbing state and receives a negative reward as shown in equation (\ref{reward}) and figure (2). Thus, the  MDP $\widetilde{M}$ gives larger penalties for taking backup safe state-actions than for going into $S_{unsafe}$. This design ensures that any nearly-optimal policy of $\widetilde{M}$ will (when running in $M$) have a high reward and low probability of visiting safety violations state-actions. 
%As we will see, as long as the safeguard policy provides safe  policies, solving the $\widetilde{M}$ will lead to a safe policy with potentially good performance in the original MDP $M$ even after we lift the safeguard module.
\begin{equation}
\label{reward}
    \widetilde{r}(s,a)= 
    \begin{cases}
    b,              & (s,a)\in \mathcal{T}_{unsafe}^{\pi}\\
    0,              & s \  \text{is an absorbing state} \\
    r,              & \text(otherwise).
    \end{cases}
\end{equation}
where $b\leq 0 $ is independent non-positive constant.

\vspace{-2mm}
\section{Experiments evaluation}
\begin{figure}[b]
    \centering
     \includegraphics[width=6cm]{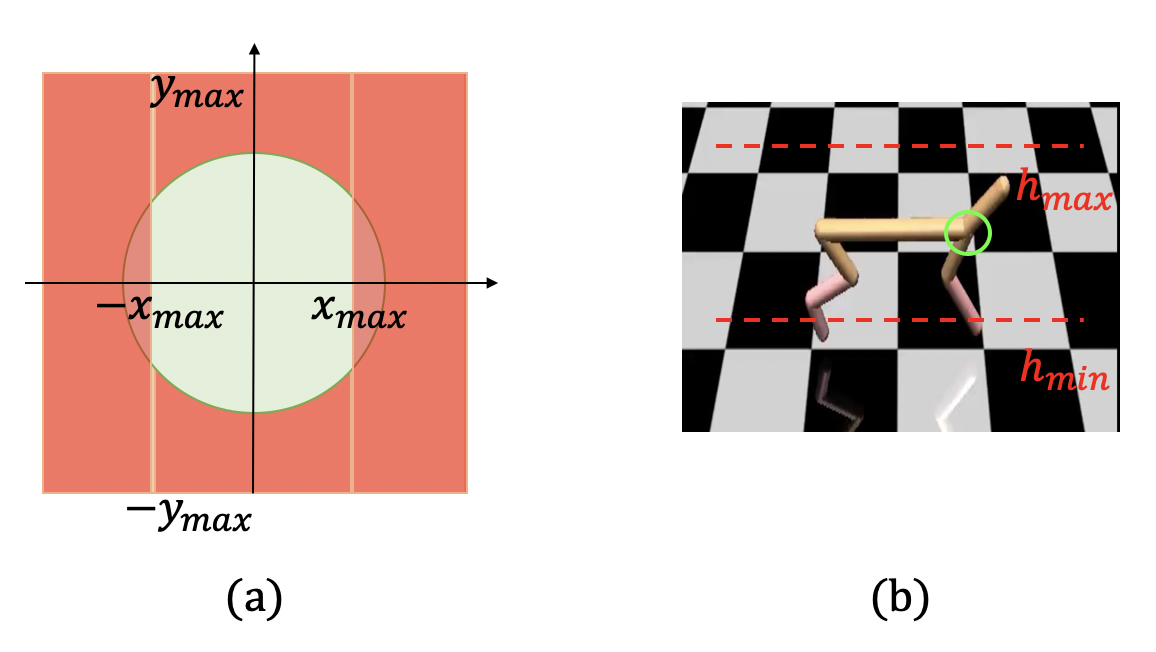}
    \caption{(a) The Circle environment. The agent can run in the green space. The green circle is the desired path, and the red lines are the constraints on the horizontal position. The vertical constraints are outside of the visualized environment. (b) The Half-cheetah environment. The green circle is centred on the link of interest, and the red dashed lines denote the allowed height range of the link.}
    \label{fig:exp_env}
\end{figure}
\vspace{-2mm}
\subsection{Experiments setup}
\vspace{-2mm}
To showcase the effectiveness of the proposed method, we test its performance on two different simulated environments, i.e.,  Circle and Half-cheetah ( figure \ref{fig:exp_env}).  To ensure fair comparisons and reproducibility of experiments, we followed the guidelines introduced by \cite{franccois2018introduction} for conducting and evaluating all of our experiments. 

\textbf{Baseline  methods}: We implement AASC by using PPO (\cite{schulman2017proximal}) as the base RL model. To complete the experiments in a reasonable amount of time, we set the number of source policies to be $I= 4$ unless mentioned otherwise. The source policies are randomly sampled from the source policy candidates. See  appendix \ref{im_detils} for all the implementation details.
% \textcolor{red}{You may need to present implementation details or mention where to find these details, as well as your implementation}
We also compared our AASC to the standard multi-layer perceptron (\textbf{MLP}) trained from scratch, which is typically used in RL literature \cite{franccois2018introduction}. As another baseline, we use  \textbf{MULTIPOLAR} \cite{barekatain2019multipolar}  which selects source policies through an adjustable matrix. Also, we consider multi-source policy reuse framework \textbf{CARL} \cite{zhang2020cautious}. 
In all the experiments, source policies are the same for AASC, MULTIPOLAR  and CARL to ensure an unbiased evaluation.
%We stress here that the existing transfer RL or meta RL approaches that train a universal policy network agnostic to the environmental dynamics, such as \cite{frans2017meta}, cannot be used as a baseline since they require a policy to be trained on a distribution of environment instances, which is not possible in our problem setting.
The following
%that ignore constraints (\textbf{Unconstrained}),  
CMDP-based approach which enforce constraints via the optimization objective \textbf{CPO} \cite{achiam2017constrained} is also considered.
%The unconstrained method optimizes task reward, ignoring constraints. 
And the \textbf{Recovery RL} \cite{thananjeyan2021recovery} which uses ideas from offline RL to pretrain the recovery policy and designs a recovery rule directly based on Q-based functions is also compared. 
% Finally, all the hyperparameters used for experiments on each environment can be found in our codebase, which will be available on the project website. 

\textbf{Environment}:
\textbf{Circle}: The circle environment ( figure \ref{fig:exp_env} (a)) is the point environment from \cite{achiam2017constrained}.  The agent is rewarded for running in a wide circle but is constrained to stay within a safe region smaller than the radius of the target circle while remaining in a circular path at high speed.
The agent has mass $m$ and can achieve maximum speed $v_{max}$. The safe set to staying within desired positional bounds $x_{max}$ and $y_{max}$, as shown in the green space in the left of figure \ref{fig:exp_env}:
$S_{safe} =\{s\in S : |x|\leq x_{max} \ \mathrm{and} \ |y|\leq y_{max}\}$
The backup policy $\mu$ applies a decelerating force (with component-wise magnitude up to $a_{max}$) until the agent has zero velocity. \textbf{Half-Cheetah}: The Half-Cheetah environment ( figure \ref{fig:exp_env} (b)) comes from OpenAI Gym and has a reward equal to the agent’s forward velocity. One of the agent’s links (green circle in the right of figure \ref{fig:exp_env}:) is constrained to lie in a given height range, outside of which the robot is deemed to have fallen over. In other words, if $h$ is the height of the link of interest, $h_{min}$ is the minimum height, and $h_{max}$ is the maximum height, the safe set is defined as $S_{safe} = \{s \in S : h_{min} \leq h \leq h_{max}\}$. The agent gets a reward equal to its forward velocity, with one of its links constrained to remain in a given height range, outside of which the robot is deemed to be unsafe.

\textbf{Evaluation metric}:
Following the guidelines of \cite{franccois2018introduction}, to measure the sampling efficiency of training policies, i.e., how quickly the training progresses, we used the average episodic reward over training samples. Furthermore, we also report the cumulative constraint violations to show the safety performance of the proposed method. We tune all algorithms to maximize the total return to see the safety performance. Each run for simulation experiments is replicated across 5 random seeds and we report the mean and standard error.

\subsection{Evaluation results}
\begin{figure}[t]
    \centering
    \includegraphics[width=11cm]{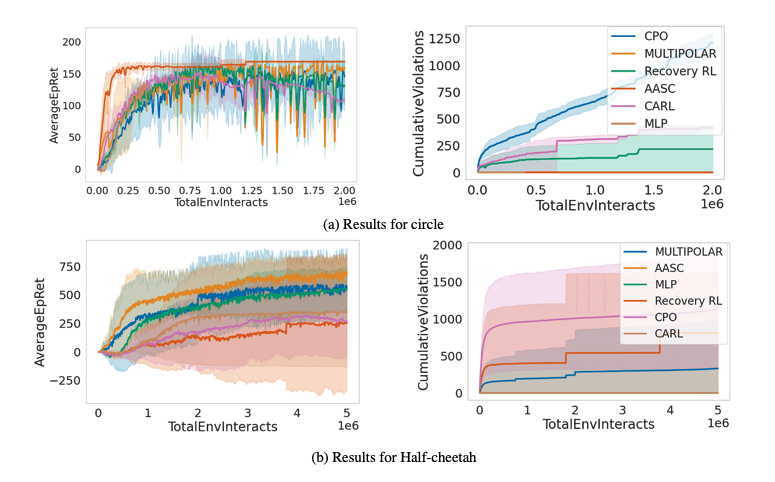}
    \caption{Results of MLP, MULTIPOLAR, CARL, CPO, Recovery RL and AASC over all the experiments for each environment. Overall AASC dramatically reduces the amount of training time and safety constraint violations while still having large returns at deployment. Plots in a row share the same legend. All error bars are ±1 standard deviation over 5 random seeds. Any curve not plotted in the third column corresponds to zero safety violations.}
    \label{fig:all_result}
\end{figure}

\vspace{-2mm}
We study the learning and safety performance of the AASC and prior methods in all simulation domains in figure 4. Results suggest that AASC significantly improve the learning efficiency remain fever safety violations than prior algorithms across two environments (Circle and Half-cheetah), which is of consistency with the motivation of our algorithm. 
% The main goal of these experiments is to test the consistency of results with the algorithm motivation.
%
%\textbf{Learning Efficiency}. 
The left column of figure \ref{fig:all_result} clearly shows that on average, AASC outperformed baseline policies in terms of sample efficiency and sometimes the final episodic reward. 
For example, the AASC converges faster at the early training stage than MLP and CPO in both environments, which indicates the effectiveness of leveraging multiple source policies in adaptive aggregation module. 
Compared with transferred RL methods such as CARL and MULTIPOLAR, AASC has always been on par or better performance on the sample efficiency. Because the aggregation module in AASC avoids the estimation of model error and leverages the change of the environment states as input, and can flexibly aggregate each action of each source policy.

%\textbf{Safety performance} 
The right column of figure \ref{fig:all_result} illustrates that the AASC prevents many safety violations. The safeguard in AASC is an unconstrained approach and allows for reliable convergence, as opposed to the baselines which rely on elaborate constrained approaches like CPO. AASC also incurs orders of magnitude fewer safety violations than Recovery RL and CARL, since the advantage-based safety evaluation is used and no model estimation error is accumulated, thus voiding the overly conservative safe region in exploration. More ablation studies could be found in appendix \ref{ablation}, where we conduct a detailed analysis of factors that affect the performance of AASC, and demonstrate that the AASC is an effective and generic framework for safe RL.
% \textcolor{blue}{Compared with recovery RL, the AASC converges fast and has fewer safety violations.}

\vspace{-2mm}
\section{Related work}
\vspace{-2mm}
\subsection{Safe reinforcement learning}
\vspace{-2mm}
% \cite{wachi2021safe}
% \cite{SafeNearFuture2021}
% \cite{cowen2022samba}
% \cite{paternain2022safe}
% Safety is one of the major concerns in the real-world applications of RL  \cite{altman1999constrained}.
Many safe RL works on endowing RL agents with the ability to satisfy constraints from the line of control theory-based and the constrained policy optimization formulation approaches.
% \cite{chow2018lyapunov,alshiekh2018safe,lutjens2019safe,cheng2019end,thananjeyan2021recovery, brown2020safe,laroche2019safe,chen2021context, zeng2021safety, luo2021learning}. 
A recent line of works on safe RL utilizes ideas from control theory and model-based approach \cite{cheng2019end,zeng2021safety,luo2021learning, chow2018lyapunov,SafeNearFuture2021,cowen2022samba}. These works propose sufficient conditions involving certain Lyapunov functions or control barrier functions that can certify the safety of a subset of states or policies. \cite{chow2018lyapunov} constructs sets of stabilizing actions using a Lyapunov function and projects the action to the set. \cite{cheng2019end} uses a barrier function to safeguard exploration and uses a reinforcement learning algorithm to learn a policy. Then \cite{luo2021learning} learns a barrier function to substitute the handcrafts 
one in \cite{cheng2019end}. From which the agent not only finds a high-return policy but also avoids undesirable states as much as possible, even during training.
 However, the works on Lyapunov functions require the discretizing of the state space and thus only work for low-dimensional space. And the barrier function-based method suffers from the sample efficiency and the bias of learned dynamics model problems \cite{SafeNearFuture2021,cowen2022samba}.

%For example, in an autonomous driving task, we do not want the robot to crash over and risk damaging itself either during training or deployment. The reason is that the agent is not able to collect experience or improve its policy until the consequences of this violation are rectified. Thus, endowing RL agents with the ability to satisfy constraints in both training and deploying not only enables robots to interact safely but also allows them to more efficiently learn in the real world. 

Another line of works design the actor-critic based algorithms under the constrained policy optimization formulation. 
% \cite{alshiekh2018safe,thananjeyan2021recovery,dalal2018safe,geibel2005risk,tessler2018reward,bharadhwaj2020conservative,achiam2017constrained,wachi2021safe,SafeNearFuture2021}. 
This line could also be divided into two groups: jointly optimizing the task performance and safety
% \cite{geibel2005risk,tessler2018reward,achiam2017constrained,dalal2018safe},  
and restricting exploration with an auxiliary policy. 
% \cite{dalal2018safe,thananjeyan2021recovery,bharadhwaj2020conservative}. 
\cite{geibel2005risk} uses a Lagrangian method to solve CMDP, and the Lagrangian multiplier is controlled adaptively \cite{tessler2018reward}. 
% \cite{achiam2017constrained} builds a trust-region around the current policy.
%
\cite{paternain2022safe} uses the first-order primal-dual optimization to solve a stochastic nonconvex saddle-point problem in CMDP, but such approaches have no guarantees of policy safety during learning. \cite{dalal2018safe} adds an additional layer, which corrects the output of the policy locally.
% They significantly reduce the number of training-time safety violations. 
However, 
% due to the natural conflict that learning new tasks in uncertain environments requires extensive exploration,  but safety requires limiting exploration, 
these approaches make safe and optimal performance trade-offs. Then \cite{bharadhwaj2020conservative} learns a conservative safety critic who underestimates how safe the policy is, and uses the conservative safety critics for safe exploration and policy optimization. \cite{thananjeyan2021recovery} makes use of existing offline data and co-trains a conservative recovery policy based on the cost-based value function. In \cite{wachi2021safe}, the authors utilize the sensors data to make the features states available, and use linear function approximation to ensure the safety performance in the discrete control environment. 
% \cite{turchetta2020safe} introduces a learnable teacher. which keeps the student safe and helps the student learn faster in a curriculum manner. However, the proposed algorithm, CISR, is still based on calling CMDP subroutines and how the intervention rules can be constructed is still not specified.
% \vspace{-1mm}
\subsection{Transfer reinforcement learning}
\vspace{-1mm}
Transfer RL aims at improving the  learning  efficiency  of  an  agent  by exploiting  knowledge  from  other  source  agents trained  on  relevant tasks. 
% It has become an important direction and a wide variety of methods have been studied in this field \cite{franccois2018introduction}. \cite{brys2015policy} applies a reward shaping approach to policy transfer, benefiting from the theoretical guarantees of reward shaping. However, it may suffer from the negative transfer. 
\cite{song2016measuring} transfers the action-value functions of the source tasks to the target task according to a task similarity metric to compute the task distance. However, they assume a well-estimated model which is not always available in practice. Later, \cite{laroche2017transfer} reuses the experience instances of a source task to estimate the reward function of the target task. 
% The limitation of this approach resides in the restrictive assumption that all the tasks share the same transition dynamics and differ only in the reward function.
%
% Besides, policy reuse is a technique to accelerate RL with guidance from previously learned policies, assuming to start with a set of available policies and select among them when faced with a new task, which is, in essence, a transfer learning approach \cite{taylor2007transfer}. 
\cite{fernandez2006probabilistic} uses the policy reuse method as a probabilistic bias when learning the new, similar tasks. \cite{rajendran2015attend} proposes the A2T (Attend, Adapt and Transfer) architecture to select and transfer from multiple source tasks by incorporating an attention network that learns the weights of several source policies for combination. 
% However, their methods select the source policy according to the performance of source policies on the target task, i.e., the utility, which fails to address the problems where multiple source policies are partially useful for learning the target task. 
\cite{barekatain2019multipolar}  uses an adjustable matrix, named MULTIPOLAR, to flexibly utilize the source policies, but omits the influence of the target environment on the selection of source solutions. 
% What's more, they all neglect the safety performance in the deployment setting which incurs the potential costly failures.
% In safety-critical control, the artificial agents can develop the ability to fast adapt and quickly learn in a new environment setting if they leverage the previously trained policy. 
Currently, there are only a few works considering leveraging learned policy \cite{zhang2020cautious,chen2021context} in safety-critical control. \cite{zhang2020cautious} employs model-based RL named CARL to train a probabilistic model to capture uncertainty about transition dynamics and catastrophic states across varied source environments. 
% It reuses the policy and all the past datasets and then fine-tunes a new task to gain good performance. %
\cite{chen2021context} proposes the context-aware safe reinforcement learning method as a  meta-learning framework to realize safe adaptation in non-stationary environments, which rely on a history of individual experiences.
% However, we cannot rely on a history of their individual experiences, as they may be unavailable due to a lack of communication between factories or prohibitively large dataset sizes. 
% For safety concerns, the current method could also not ensure the none safety violations because exploring the environment to learn about constraints requires a significant amount of constraints violations during learning.

\vspace{-2mm}
\section{Conclusion}
\vspace{-2mm}
%Instead of solving (2) as a CMDP problem as a whole problem, the proposed method separates the often conflicting objectives of task-directed optimization and constraint satisfaction allowing it to not only be safer during learning but also learn more efficiently. Compared with the original $M$, the proposed method using shield policy constructs the new MDP and has more states which can lead to absorbing state-action pairs and assigns lower rewards to them, which capture state $S_\triangleright$.
%Transferring knowledge from prior source tasks in solving a new target task can be useful in several learning applications to improve the data efficiency of safe RL problems.
% We consider the problem of safe exploration in reinforcement learning, where the goal is to discover a policy that maximizes the expected return, but additionally, desire the training process to incur minimal safety violations.  
In this work, we propose an adaptive aggregation framework for safety-critical control, which aims to improve sample efficiency and safety performance.  We first learn to aggregate the safe actions provided by the source policies adaptively to maximize the target task performance.  Meanwhile, we learn an auxiliary network that predicts residuals around the aggregated safe actions, which ensures the target policy’s expressiveness even when some of the source policies perform poorly. What's more, we separate the constraints and explorations and use an advantaged safeguard evaluation to ensure safety during the learning process. Separating the task and safeguard policies makes it easier to balance task performance and safety, and allows using off-the-shelf RL algorithms for both.  Empirically, our algorithm compares favourably to state-of-the-art safe RL methods, in terms of the trade-off between the learning and safety performance, and can achieve higher sample efficiency.
\bibliographystyle{unsrtnat}
\bibliography{my_bib.bib}
%%%%%%%%%%%%%%%%%%%%%%%%%%%%%%%%%%%%%%%%%%%%%%%%%%%%%%%%%%%

\newpage
\appendix
% \input{tex/theory}
% \section{Algorithm details}
\section{Implementation details}
\label{im_detils}

\subsection{Network architectures}
We implement baselines and AASC by using PPO (Schulman et al., 2017) as the base RL model.  The \textbf{MLP} is a special case of AASC without source policies, which means training from scratch. The MLP of the Circle environment consists of 2 hidden layers, 64 neurons per hidden layer, tanh activation function, with batch size 4000 and discount factor $\gamma$ 0.99. The threshold $\sigma$ is 0.01 and penalty value $b$ is -2 and cost constant $\alpha$ is 0.5. The advantage threshold $\eta=0.08$. Both network parameters are updated using Adam %\cite{kingma2014adam} 
with a learning rate of $10^{-3}$. All target networks are updated with a learning rate of $10^{-3}$. For the Half-cheetah environment, the network design is the same as the Circle environment and the penalty value $b$ is -0.01 and the cost constant $\alpha$ is 0.05 and $\eta=0.2$. The same network architecture is also used as a base model in others baselines unless specifically stated. 

In \textbf{AASC}, the auxiliary network has identical architecture to that of the MLP. Therefore, the only difference between MLP and AASC was the adaptive aggregation part, which makes it possible to evaluate the contribution of transfer learning based on an adaptive aggregation of source policies. The attention network in AASC also has the same architecture as MLP, with the difference in the output layer with softmax function.

For a fair comparison, we also implement the \textbf{MULTIPOLAR} with the safeguard evaluation module to illustrate the advantages of using the adaptive aggregation of sources policy. The MULTIPOLAR selects source policies through an adjustable matrix. Since MULTIPOLAR does not utilize the effects of the input states to adjust the differentiable matrix directly, its ability to adjust the weighted parameters according to the change in the environment is weaker than AASC.

In \textbf{Recovery RL}, the unsafe set is defined as $\mathcal{T}_{unsafe}^{\pi} = \{(s,a)\in S_{safe}\times A:Q(s,a_{task})\geq\eta\}$, where
$\mathcal{T}_{safe}^{\pi} = S\times A  \setminus \mathcal{T}_{unsafe}^{\pi}$. The value of $\eta$ is same as in AASC. Plus, the Recovery RL trains the agent with a set of transitions $D_{offline}$ that contain constraint violations for pretraining. Since we don't have the prior knowledge of the target environment, we utilize this pretrained dataset and fine-tune the network in the target task.

In \textbf{CARL}, we employ model-based RL to train a probabilistic model to capture uncertainty about transition dynamics and catastrophic states across varied source environments. Then fine-tune on a new task. The parameters are chosen with little modification to the original
training parameters of the similar environments in \cite{zhang2020cautious}.

In \textbf{CPO}, we guarantee safety in terms of constraint satisfaction that holds in expectation and extends trust-region policy optimization (TRPO) to handle the CMDP constraints, the parameters are the same as in \cite{achiam2017constrained}.

All experiments were run on an NVIDIA GeForce GTX 1070. 
The given hyperparameters were found by hand-tuning until the good performance was found on all algorithms.

\subsection{Environment parameters}
% \textbf{Circle environment}:
The state in Circle environment can be represented as  $s = (x, y, \dot{x}, \dot{y})$, where $(x, y)$ is the x-y position and $(\dot{x}, \dot{y})$ is the corresponding velocity. The action $a = (a_x, a_y)$ is the force applied to the robot (each component has maximum magnitude $a_{max}$). 
In our experiments, we set the parameters of target environment to $m = 1, v_{max} = 2, a_{max} = 1,  x_{max} = 2.5$, and $y_{max} = 15$. In Half-cheetah environment, we set $h_{min} = 0.4$ and $h_{max} = 1$. 

In all experiments, when computing $\overline{Q}$, we use the cost function which indicates the distance to the unsafe set to make our intervention mechanism reasonable and hence the training process safer. $\hat{c}(s,a)$ is a function of the distance of the state $s$ to the boundary of the unsafe set, denoted by $\mathrm{dist}(s, S_{unsafe})$. For the circle environment, $S_{unsafe}$ denotes the unsafe regions which are outside the vertical lines. The distance function is $\mathrm{dist}(s, S_{unsafe}) = \mathrm{max}\{0, \mathrm{min}\{x_{max}-x, x_{max}+x, y_{max}-y, y_{max}+y\}\}$. For the half-cheetah environment, the distance function $\mathrm{dist}(s, S_{unsafe}) = \mathrm{max}\{0, \mathrm{min}\{h - h_{min}, h_{max} - h\}\}$. For some constant $\alpha \geq 0$, the cost function is defined as a hinge function of the distance as:
\begin{equation}
    \hat{c}(s,a)=
    \begin{cases}
    \mathbbm{1}\{\mathrm{dist}(s, S_{unsafe})=0\},              & \alpha = 0\\
    \mathrm{max}\{0, 1-\frac{1}{\alpha}\mathrm{dist}(s, S_{unsafe})\},              & \text(otherwise).
    \end{cases}
\end{equation}

We note that $\hat{c}$ is an upper bound for $c$ if $\alpha > 0$ and $\hat{c} = c$ if $\alpha = 0$. This cost function is continuous so that the effects of approximation bias are smaller than that resulting from a binary cost ${0,1}$. The backup policy $\mu$ applies a decelerating force (with component-wise magnitude up to $a_{max}$) until the agent has zero velocity. 

\begin{figure}
  % \centering
  \includegraphics[width=0.8\linewidth]{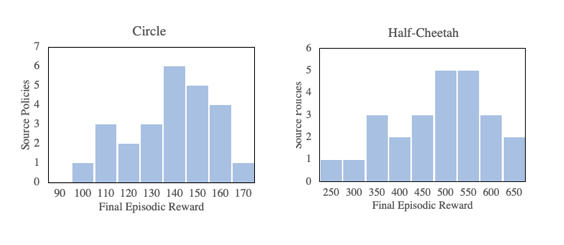}
  \caption{Histogram of final episodic rewards obtained by source policies in their original environment instances.}
  \label{fig_hisograms}
\end{figure}

\subsection{Source policies}
For each of the environments,  we first created  25  environment instances by randomly sampling the dynamics and kinematics parameters from a specific range.  For example, these parameters in the Circle environment were the agent's mass $m$, constraint width $x_{max}$ and length $y_{max}$, max velocity $v_{max}$ defined in the environment and height constraints $h_{max}$ and $h_{min}$ in  Halfcheetah. Details of sampling ranges for parameters of all environments are presented in table \ref{tab:tab1} and \ref{tab:tab2}. Note that we defined the sampling ranges for each environment such that the resulting environment instances are significantly different in dynamics, as shown in figure 1.  Figure \ref{fig_hisograms} shows the histograms of final episodic rewards (average rewards of the last 100 training episodes) for the source policy candidates obtained in their original environment instances. As shown in figure 1, the source policies were diverse in terms of performance. Then, for each environment instance, we trained MLP policies as the source policy candidates, from which we sample $I=4$ of them to train AASC as well as CARL and MULTIPOLAR. Specifically, for each environment instance, we trained three sets of policies each with distinct source policy sets selected randomly from the candidate pool. Plus, the learning procedure explained above was done three times with fixed different random seeds to reduce variance in results due to stochasticity. What's more, although we used trained MLPs as source policies for reducing experiment times, any type of policy including hand-engineered ones could be used for AASC in principle.

% In all the transfer learning-based safe RL like CARL and MULTIPOLAR experiments, source policies are the same  to ensure an  unbiased  evaluation.  
%  which is differentiable with dimension $(I+1) * D$, and $I=4$
\vspace{5mm}
\begin{table}[h]
    \caption{Sampling ranges for Circle environment parameters}
    \vspace{2mm}
    \centering
    \begin{tabular}{cc}
    \toprule
    Factors & Value range \\
    \midrule
    $m$ &  [0.5, 1.5]\\
    $x_{max}$ & [0.5,  3.5] \\
    $y_{max}$ &  [5, 25]\\
    $v_{max}$ & [1, 3] \\
    \bottomrule
    \end{tabular}
    \label{tab:tab1}
\end{table}

\vspace{10mm}
\begin{table}[h]
    \caption{Sampling ranges for Half-cheetah environment parameters}
    \vspace{2mm}
    \centering
    \begin{tabular}{cc}
    \toprule
    Factors & Value range \\
    \midrule
    $m$ & [10, 18] \\
    $h_{max}$ & [0.8, 1.2] \\
    $h_{min}$ & [0.2, 0.6]\\
    % $v_{max}$ & [1,3] \\
    \bottomrule
    \end{tabular}
    \label{tab:tab2}
\end{table}

% Our experiments consider the following approaches to construct $\Bar{Q}$.

\section{Ablation Study}
\label{ablation}

\textbf{Effect of source policy performances}.

 In this setup, we investigate the effect of source policy performances on AASC sample efficiency. We select two separate pools of source policies, where one contained only high-performing and the other only low-performing ones. For example, in the Circle environment, the high-performing ones are those with a final episodic reward higher than 150, and the low-performing ones are those under 120. Table \ref{tab:tab3} summarizes the results of sampling source policies from these pools (4 high, 2 high \& 2 low, and 4 low performances) and compares them to the original AASC after 250k and 500k training steps. As shown in this table, AASC can achieve the best performance when all the source policies were sampled from the high-performance pool. However, we emphasize that such high-quality policies are not always available in practice, due to the variability of how they are learned or hand-crafted in their environment instance. Interestingly, by comparing the reported results in table \ref{tab:tab3} MLP and AASC with 4 low performance, we can observe that even in the worst-case scenario of having only low performing source policies, the sample efficiency of AASC is on par with that of learning from scratch. This suggests that AASC avoids the worse source policy transfer, which occurs when transfer degrades the learning efficiency instead of helping it. 
 Further, AASC successfully learns to suppress the useless low-performing sources to maximize the expected return in a target task, indicating the mechanism of source policy rejection.

\vspace{5mm}
\begin{table}[h]
    \caption{Results with different source policy sampling in Circle environment}
    \vspace{2mm}
    \centering
    \begin{tabular}{ccc }
    \toprule
    Methods & 250k & 500k\\
    \midrule
    MLP &  86 (80, 94) & 125 (117, 138)\\
    % MULTIPOLAR & 139(130,140) & \\
    % CARL &  135(127,140) &\\
    AASC (random)& 135 (131, 147) & 145 (135, 157)\\
    AASC (4 high performance) & 156 (152, 160) & 158 (152, 163)\\
    AASC (2 high \& 2 low performance) & 132 (121, 137)& 142 (138, 154)\\
    AASC (4 low performance) & 85 (84, 98) &  126 ( 119, 140)\\
    \bottomrule
    \end{tabular}
    \label{tab:tab3}
\end{table}
\vspace{10mm}

\textbf{Effect of number of source policies}. 

Besides, we show how the number of source policies contributes to AASC’s sample efficiency in Table \ref{tab:tab4}. Specifically, we trained AASC policies up to $K = 10$ to study how the mean of average episodic rewards changes. The monotonic performance improvement over $K$ for $K\leq10$ is achieved at the cost of increased inference time. The AASC with $K=4$ can achieve a competitive performance compared with $K=10$ considering the computation complexity and the availability of source policies. In practice, we suggest balancing this speed-performance trade-off by using as many source policies as possible before reaching the inference time limit required by the application.

\vspace{5mm}
\begin{table}[h]
    \caption{Results with different number of source policies in Circle environment}
    \vspace{2mm}
    \centering
    \begin{tabular}{ccc }
    \toprule
    Methods & 250k & 500k\\
    \midrule
    MLP &  86 (80, 94) & 125 (117, 138)\\
    % MULTIPOLAR & [0.5,  3.5] & \\
    % CARL &  [5, 25] &\\
    AASC $(K=1)$ &  127 (125, 135) & 138 (136, 142)\\
    AASC $(K=4)$ & 135 (131, 147) & 145 (135, 157)\\
    AASC $(K=10)$ & 137 (133, 150) & 150 (140, 160)\\
    \bottomrule
    \end{tabular}
    \label{tab:tab4}
\end{table}
\vspace{10mm}

% \textbf{Effect of Adaptive Aggregation of Safe Policies}.

% To demonstrate the importance of each component of AASC, we also evaluated the following degraded versions: (1) $\pi_{aux}$ fixed to 1. (2) attention networks fixed to $\frac{1}{(I+1)\times T}$, which just averages the actions from the source policies without adaptive weights (similar to the residual policy learning methods that use raw action outputs of a source policy), and (3) $\pi_{aux}$ fixed to 0.  The full version of 
% AASC significantly outperformed the degraded ones, suggesting that adaptive aggregation and predicting $a_{aux}$ are both critical.

% \textbf{Effect of Safeguard Evaluation}.

% We omit the safeguard module and only preserve the adaptive aggregation of safety policies.
% % \textbf{Effect of Varying Intervention Penalty}.

% % We vary the intervention penalty $b$ for the Circle environment, the results \textcolor{blue}{should be consistent with the performance and safety bounds in Theorem 1}

% We verify our proposed method can drastically reduce the adaptation step and the amount of the unsafe trajectories generated in training in the new experiment setting, while still resulting in good safety and performance in deployment. 

\end{document}